\documentclass[letterpaper, 10 pt, conference]{ieeeconf}  

\usepackage{nicefrac}
\usepackage{siunitx}
\usepackage{svg}
\usepackage{booktabs}
\usepackage{array,framed}
\usepackage{booktabs}
\usepackage{
  color,
  float,
  epsfig,
  wrapfig,
  graphics,
  graphicx,
  subcaption
}
\usepackage{textcomp,amssymb}
\usepackage{setspace}
\usepackage{latexsym,fancyhdr,url}
\usepackage{enumerate}
\usepackage{algorithm2e}
\usepackage{algpseudocode}
\usepackage{graphics}
\usepackage{xparse} 
\usepackage{xspace}
\usepackage{multirow}
\usepackage{csvsimple}
\usepackage{balance}

\usepackage{
  tikz,
  pgfplots,
  pgfplotstable
}
\usepackage{hyperref}

\usetikzlibrary{
  shapes.geometric,
  arrows,
  external,
  pgfplots.groupplots,
  matrix
}

\pgfplotsset{compat=1.9}

\usepackage{mathtools,}

\IEEEoverridecommandlockouts                              

\title{\LARGE \bf
Attention-based Learning for 3D Informative Path Planning
}

\author{Rui Zhao$^{1}$, Xingjian Zhang$^{1}$, Yuhong Cao$^{1}$, Yizhuo Wang$^{1}$, Guillaume Sartoretti$^{1}$ 
\thanks{$^{1}$Authors are with the Department of Mechanical Engineering, College of Design and Engineering, National University of Singapore
        {\tt\small zhao\_rui1998@outlook.com,\{xingjian,caoyuhong,
        wy98\}@u.nus.edu,mpegas@nus.edu.sg}}%
}

\begin{document}
\maketitle
\thispagestyle{empty}
\pagestyle{empty}
\begin{abstract}

In this work, we propose an attention-based deep reinforcement learning approach to address the adaptive informative path planning (IPP) problem in 3D space, where an aerial robot equipped with a downward-facing sensor must dynamically adjust its 3D position to balance sensing footprint and accuracy, and finally obtain a high-quality belief of an underlying field of interest over a given domain (e.g., presence of specific plants, hazardous gas, geological structures, etc.). In adaptive IPP tasks, the agent is tasked with maximizing information collected under time/distance constraints, continuously adapting its path based on newly acquired sensor data. To this end, we leverage attention mechanisms for their strong ability to capture global spatial dependencies across large action spaces, allowing the agent to learn an implicit estimation of environmental transitions. Our model builds a contextual belief representation over the entire domain, guiding sequential movement decisions that optimize both short- and long-term search objectives. Comparative evaluations against state-of-the-art planners demonstrate that our approach significantly reduces environmental uncertainty within constrained budgets, thus allowing the agent to effectively balance exploration and exploitation. We further show our model generalizes well to environments of varying sizes, highlighting its potential for many real-world applications.

\end{abstract}


\section{Introduction}
\label{sec:intro}

Autonomous aerial robots have gained increasing traction for a wide range of sensing and monitoring tasks, including weed detection in agriculture~\cite{detweiler2015environmental,hitz2014fully}, hazardous gas detection in mines~\cite{dunbabin2012robots, neumann2012autonomous}, and geological or environmental surveying~\cite{barfoot2010field, muscato2012volcanic}. Given limited resources of a mobile robot such as battery life and mission time/distance, the robot must methodically plan its path to reconstruct the targeted field as accurately as possible within these constraints. This problem is known as the informative path planning (IPP) problem, which focuses on strategically selecting measurement locations to maximize information gain while adhering to predefined budgets or resource limits~\cite{hitz2017adaptive, popovic2017online}.

In this work, we focus specifically on \emph{adaptive} IPP, where the agent needs to continuously update its path online based on newly acquired sensor data~\cite{ruckin2022adaptive}.
Adaptive planning is particularly crucial in large-scale or previously unknown environments, where prior information is either unavailable or highly uncertain.
In such cases, adaptivity allows the robot to dynamically adjust its path in response to complex spatial correlations, hidden hotspots, or evolving field conditions, which is often beyond the capability of traditional precomputed (non-adaptive) methods.

\begin{figure}[t]
\vspace{0.1cm}
\setlength{\belowcaptionskip}{-5mm}
  \centering
    \includegraphics[width=.5\textwidth]{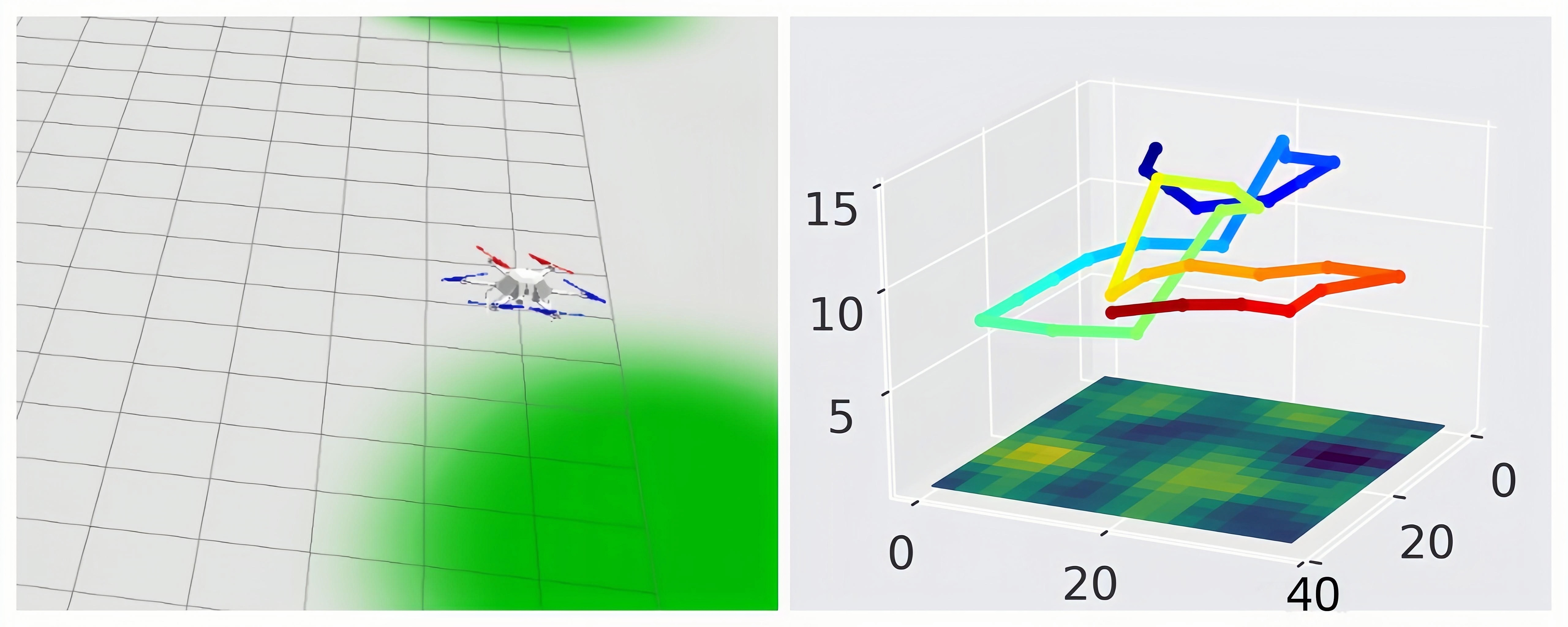}
    \vspace{-0.65cm}
    \caption{
    \textbf{Demonstration of our attention-based IPP planner in a weed-detection scenario.}
    \emph{(Left)} The UAV’s simulation environment, with scattered weed regions (green).
    \emph{(Right)} A 3D path from blue to red over time, projecting weed occupancy (lighter cells) onto the grid. The high-density areas are adaptively prioritized in our work.
    }
    \label{illustrate}
    \vspace{-0.2cm}
\end{figure}

However, recent progress in two-dimensional IPP~\cite{popovic2020informative, choudhury2020adaptive} does not fully leverage the ability of unmanned aerial vehicles (UAVs) to adjust their altitude.
Many onboard sensors, such as cameras, have a field of view (FoV) that expands as the UAV ascends, but this comes at the cost of reduced measurement accuracy.
As a result, one of the core challenges in 3D IPP stems from the combinatorial complexity introduced by the increased dimensionality of the action space and state representation.
Thus, achieving fully adaptive IPP in 3D action spaces that can dynamically balance observation accuracy and sensing footprint across different mission phases remains an open research problem for the practical deployment of UAVs in IPP tasks.


Recent advances in deep reinforcement learning (DRL) offer promising tools for addressing these long-horizon planning challenges~\cite{chen2020autonomous, viseras2019deepig}.
In particular, self-attention mechanisms~\cite{vaswani2017attention} demonstrated strong capabilities in capturing global dependencies in sequential data, making them well-suited for modeling spatial correlations in environmental fields.
Building upon this insight, we propose an attention-based IPP approach that enables an autonomous UAV to actively sense and explore an unknown environment directly in 3D action space, subject to limited travel resources.
To reduce complexity, we discretize the continuous search space into a sparse graph that spans the environment, with multiple levels corresponding to different flight altitudes.
Each node in the graph serves as a candidate decision point for the agent, and the agent navigates between these nodes while collecting measurements at a fixed frequency.
The graph is associated with the agent's \textit{belief} representation and helps us formulate adaptive 3D IPP as a sequential decision-making problem on the graph.
We train a policy network to adaptively select the next neighboring node to visit, considering both horizontal position and flight altitude, and their effect on sensing footprint and accuracy.
Our experiments show that our proposed approach yields superior performance and generalizes to environments of varying sizes, demonstrating its scalability and robustness to a wide range of scenarios.




\section{Related Work}
\label{sec:relwork}

Informative path planning problems can be formulated as partially observable Markov decision processes (POMDPs)~\cite{kaelbling1998planning}, providing a principled way for planning under uncertainty. These formulations often leverage Shannon entropy-based metrics to quantify information gain~\cite{charrow2015information,hollinger2014sampling}, yet their high dimensionality demands more computationally efficient approaches to enable real-time planning.

Research on IPP can be broadly divided into two facets: environment modeling~\cite{o2012gaussian,sun2015bayesian,vasudevan2009gaussian} and efficient data acquisition~\cite{binney2013optimizing,hitz2014fully,hollinger2014sampling}. In terms of modeling, Gaussian Processes (GPs) have emerged as a key Bayesian tool for capturing spatial–temporal correlations~\cite{seeger2004gaussian,wang2023spatio}, enabling IPP approaches to incorporate covariance structures~\cite{singh2010modeling} and handle large-scale datasets via approximation techniques~\cite{vasudevan2009gaussian}, thereby enhancing the utility of collected measurements. On the data acquisition side, Arora \emph{et al.}~\cite{arora2017randomized} proposed fusing Constraint Satisfaction Problems with the Travelling Salesman Problem to develop Randomized Anytime Orienteering (RAOr), while Hitz \emph{et al.}~\cite{hitz2017adaptive} employed the Covariance Matrix Adaptation Evolution Strategy (CMA-ES), an evolutionary algorithm, to generate high-quality candidate paths for IPP in static environments. More recent approaches, such as CAtNIPP~\cite{cao2023catnipp}, leverage context-aware attention mechanisms to learn adaptive IPP policies in 2D settings by integrating image-like sensing. However, like most of the aforementioned methods, these approaches operate solely in 2D action spaces.
Notably, one recent study explicitly addresses IPP for aerial robots in 3D action spaces~\cite{apoorva2024adaptiveipp}.
However, this work still adopts probe-based point sensing, which is unaffected by the UAV's altitude, rather than image-like sensing, where the agent must learn to reason about how its altitude affect the sensor's field of view and accuracy.

In general, IPP seeks to minimize uncertainty under budget constraints, a problem known to be NP-hard~\cite{krause2008near} or even PSPACE-hard~\cite{reif1979complexity}. Discrete IPP solutions often employ combinatorial optimization over grid representations~\cite{chekuri2005recursive,hollinger2009efficient}, whereas continuous IPP methods commonly use sampling-based planners~\cite{hollinger2014sampling} or spline parametrization~\cite{hitz2017adaptive,charrow2015information}. Despite their success, both approaches face scalability and resolution constraints, particularly in high-dimensional 3D spaces. Prior attempts to address large-scale domains include discretizing the action space via sparse graphs~\cite{choudhury2020adaptive,popovic2020informative}, though this can diminish plan accuracy. More recently, reinforcement learning (RL) has been explored to directly learn data-gathering strategies, yielding promising results~\cite{chen2020autonomous,viseras2019deepig} but primarily in low-dimensional action spaces. Extending IPP to adaptive 3D planning—especially in settings where spatial correlations and large-scale action spaces are critical—thus remains a largely unexplored challenge.


\section{Background}
\label{sec:background}

\subsection{Problem Formulation}
Following~\cite{popovic2020informative}, we consider a set of candidate measurement positions 
\(\psi = (\psi_{1}, \dots, \psi_{n})\) in \(\mathbb{R}^3\). Our goal is to maximize an information-gain metric while respecting a budget \(B\), such as a flight-time or energy limit. Formally, the problem is to find:
\begin{equation}
\psi^* = \underset{\psi \in \Psi}{\mathrm{argmax}} \; I(\psi) 
\quad \text{s.t.} \quad 
\sum_{i=1}^{n-1} c(\psi_i,\psi_{i+1}) \leq B,
\label{eq:3dipp-objective}
\end{equation}
where \(c(\psi_i,\psi_{i+1})\) represents the traversal cost (e.g., time) between consecutive measurement sites. Following~\cite{binney2012branch,popovic2020informative}, the information gain \(I(\psi)\) is defined as:
\(I(\psi) = \mathrm{Tr}(\mathbf{P}^{-}) - \mathrm{Tr}(\mathbf{P}^{+})\), 
where \(P^{-}\) and \(P^{+}\) are covariance submatrices of the environment's belief before and after collecting measurements along \(\psi\). 

Crucially, the integration of adaptivity enables the planning system to prioritize Regions of Interest (ROIs), identified based on the following criteria:
 \(\mathcal{X}_I = \{\mathbf{x}_i \mid \mathbf{x}_i \in \mathcal{X} \wedge \mu_i +\beta\sigma_i\ge \mu_{\mathrm{th}}\}\), 
where \(\mu_{\mathrm{th}}\) is a chosen threshold on the scalar field of interest, and \(\beta\) is a design parameter to weight uncertainty (i.e., \(\beta\sigma_i\)).

\subsection{Gaussian Process Representation}
To represent an underlying scalar field \(\zeta\) (e.g., weed distribution or vegetation index), we assume a Gaussian Process prior, \(\zeta \sim \mathcal{GP}(\boldsymbol\mu, \mathbf{P})\). For an environment discretized into \(n\) points \(X\) and any prediction set \(X^{*}\), the posterior covariance can be written as~\cite{seeger2004gaussian,vasudevan2009gaussian}:
\begin{equation} 
\begin{aligned}
P & = K(X^*, X^*)- K(X^*, X) 
    \big[ K(X, X) + \sigma_n^2 I \big]^{-1} \\
  &\quad \times K(X^*, X)^\top,
\end{aligned}
\label{gaussian}
\end{equation}
where \(K(\cdot,\cdot)\) denotes the covariance kernel and \(\sigma_n^2\) is sensor noise variance. In practice, we often initialize the field mean \(\boldsymbol\mu\) uniformly (e.g., 0.5 as a typical choice for $0-1$ distribution) and update it as measurements are collected.
\subsection{Sequential Data Fusion}
After each new sensor reading, the GP posterior \(\mathcal{GP}(\boldsymbol{\mu}^{+}, \mathbf{P}^{+})\) is refined through a Kalman-like update~\cite{reece2010introduction}. Let \(\mathbf{z} = [z_{1}, \dots, z_{m}]^{\top}\) be \(m\) independent measurements associated with points \(\{x_{1}, \dots, x_{m}\} \subset X\). Each measurement \(z_{i}\) has an estimated mean \(\mu_{s,i}\) and variance \(\sigma_{s,i}^{2}\). We collect them in the measurement vector \(\mathbf{z}\) and update:
\begin{equation}
\begin{aligned}
\boldsymbol{\nu}^{+} = &\; \boldsymbol{\nu}^{-}+
\mathbf{K}\mathbf{v},\\[4pt]
\mathbf{P}^{+} = &\; \mathbf{P}^{-} - \mathbf{K}\mathbf{H}\mathbf{P}^{-},
\end{aligned}
\label{eq:3dipp-kf}
\end{equation}
where \(\boldsymbol{\nu}^{-}\) and \(\boldsymbol{\nu}^{+}\) are the pre- and post-update mean vectors, respectively. The Kalman gain \(\mathbf{K} = \mathbf{P}^{-}\mathbf{H}^{\top}\mathbf{S}^{-1}\) uses the covariance innovation \(\mathbf{S} = \mathbf{H}\,\mathbf{P}^{-}\mathbf{H}^{\top} + \mathbf{R}\), and \(\mathbf{v} = \mathbf{z} - \mathbf{H}\,\boldsymbol{\nu}^{-}\). Here, \(\mathbf{H}\) projects map states onto measurements, and \(\mathbf{R}\) is a diagonal matrix capturing measurement noise. This method can become computationally expensive as \(\mathbf{P}^{-}\) grows with the map size~\cite{jin2022adaptive}, but remains tractable for moderately large environments via efficient approximations.

\subsection{Height-Dependent Sensor Model}
\begin{figure}[t]
  \centering
  \setlength{\belowcaptionskip}{-3mm}
    \includegraphics[width=0.45\textwidth]{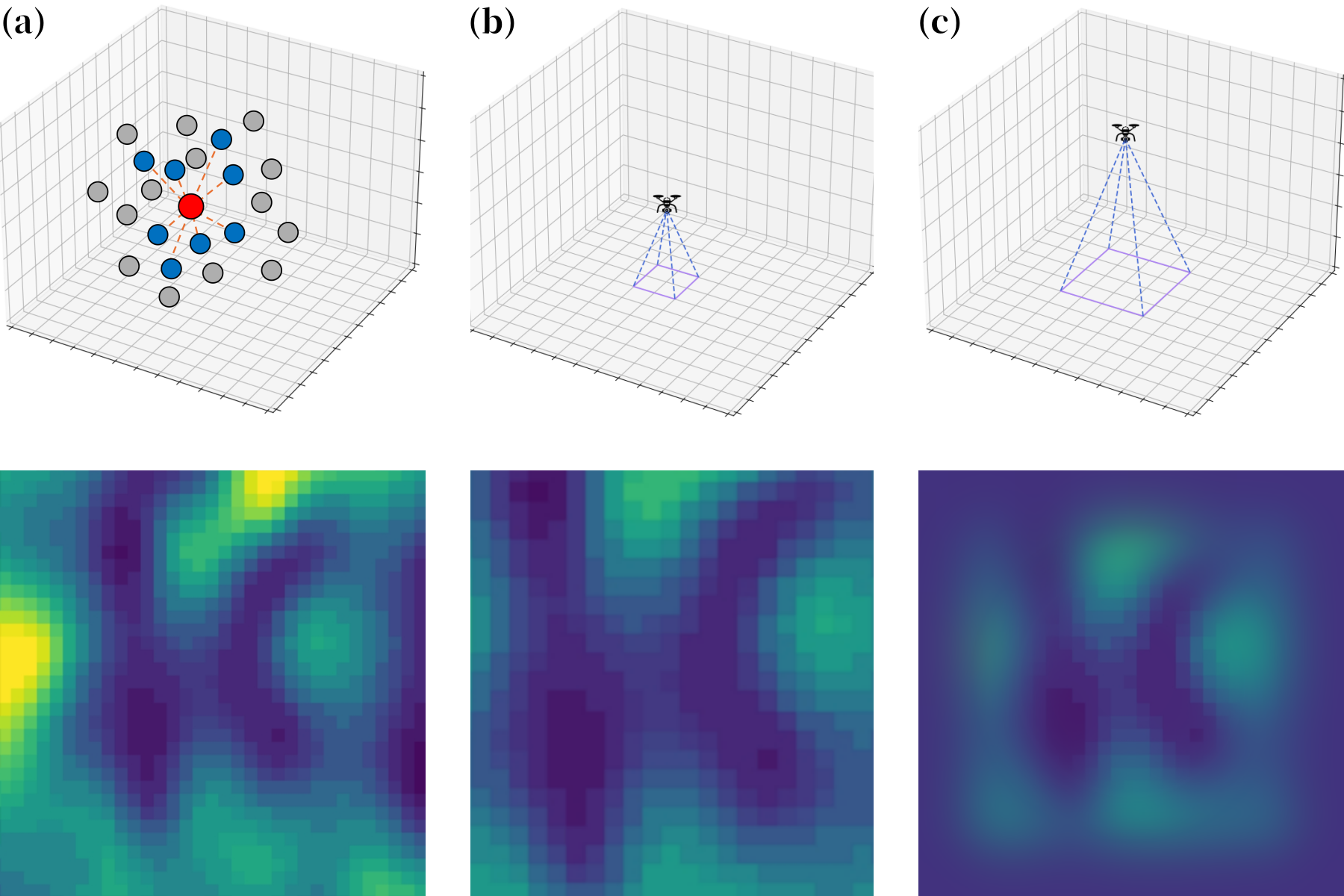}
    \caption{\textbf{Sensor model and Kalman filter data fusion with agent belief.} (a) Ground truth field distribution and the complete 3D probabilistic roadmap, where the agent’s current node (red) is connected to neighboring nodes (blue), with all other non-neighboring nodes shown in gray; (b) An independent measurement taken at $10 \rm m$ altitude, along with the corresponding belief update starting from an empty prior; (c) Another independent measurement taken at $20 \rm m$ altitude, with the corresponding empty belief update showing increased uncertainty due to the wider FoV at higher altitude.}
    \label{FoV}
\end{figure}

Following~\cite{popovic2020informative}, each sensor measurement \(z_i\) is assumed to be Gaussian-distributed, \(z_i \sim \mathcal{N}(v_{s,i},\,\sigma_{s,i}^{2})\). The variance \(\sigma_{s,i}^{2}\) grows with altitude \(h\) following 
\(\sigma_{s,i}^{2} = a\bigl(1 - e^{-bh}\bigr)\), 
where \(a\) and \(b\) are user-chosen parameters. As the UAV ascends, sensor noise increases while the FoV expands. In many applications, the FoV above a certain altitude (e.g., $15\rm m$) is scaled by a factor, effectively trading measurement breadth for reduced precision. Figure~\ref{FoV} demonstrates this trade-off: higher flight altitudes allow the platform to cover more areas with fewer stops, yet each individual observation becomes noisier. For binary mapping tasks, grid cells typically store a probability indicating the likelihood of a target (label~1) versus a non-target (label~0).

\section{Method}
\label{sec:methodology}

\subsection{RL Environment for 3D IPP}
We begin by using a probabilistic roadmap (PRM)~\cite{geraerts2004comparative} to approximate the continuous 3D workspace with a graph \(G=(V,E)\). Each node \(v_i = (x_i, y_i, z_i) \in V\) is connected to its \(k\) nearest neighbors via edges in \(E\). Next, we recast the adaptive IPP problem as a sequential decision-making process. In this framework, the agent iteratively selects which node to visit based on the current environment observations subject to $E$, thereby constructing the final path \(\phi = (\phi_0, \phi_1, \dots, \phi_n)\). Below, we detail how observation, action, and reward are specified within this RL formulation.

\subsubsection{Observation}
At time step \(t\), the agent's observation \(s_t\) is defined as $s_t = \{G',\, v_c\}$, where \(G' = (V', E)\) is the \emph{augmented} graph. In \(G'\), each node \(v'_i \in V'\) stores both the original coordinates \(\bigl(x_i, y_i, z_i\bigr)\) and the associated Gaussian Process  estimates, \(\boldsymbol\mu\bigl(v_i\bigr)\) and \(\mathbf{P}\bigl(v_i\bigr)\). These represent the mean and uncertainty of the environment in the sensor's FoV at node \(v_i\). The notation \(v_c\) designates the agent's current node. For better generalization to diverse environments, node coordinates are normalized:
\begin{equation}
d' \;=\; \frac{\,d - \min(d)\,}{\,\max(d) - \min(d)\,},
\label{norm}
\end{equation}
where \(d\) can be \(x\), \(y\), or \(z\)-coordinate values.

\subsubsection{Action}
At each decision step \(t\), the agent chooses its next measurement node from the \(k\) neighboring vertices of the current node \(v_c\). Formally, the policy \(\,p_{\theta}\bigl(\phi_{t}\bigr)\) is a probability distribution over these neighbors, parameterized by \(\theta\): $p_{\theta}\bigl(\phi_t = v_i \mid s_t\bigr),
\quad
\text{where}\;
(v_c,v_i)\in E.$
Once the agent selects a node \(v_i\), it travels in a straight line from \(v_c\) to \(v_i\), thereby extending its overall path \(\phi\). To ensure a balanced distribution of neighboring nodes across all altitude levels, we implement a selection strategy that incorporates $k/n$ points at each altitude, where $n$ represents the total number of distinct altitude levels. This method enables the agent to explore a broader spectrum of altitudes when selecting the next observation location, promoting more informed decision-making.


\subsection{Network Structure}
\begin{figure*}[ht]
\setlength{\belowcaptionskip}{-4mm}
  \centering
    \includegraphics[width=.85\textwidth]{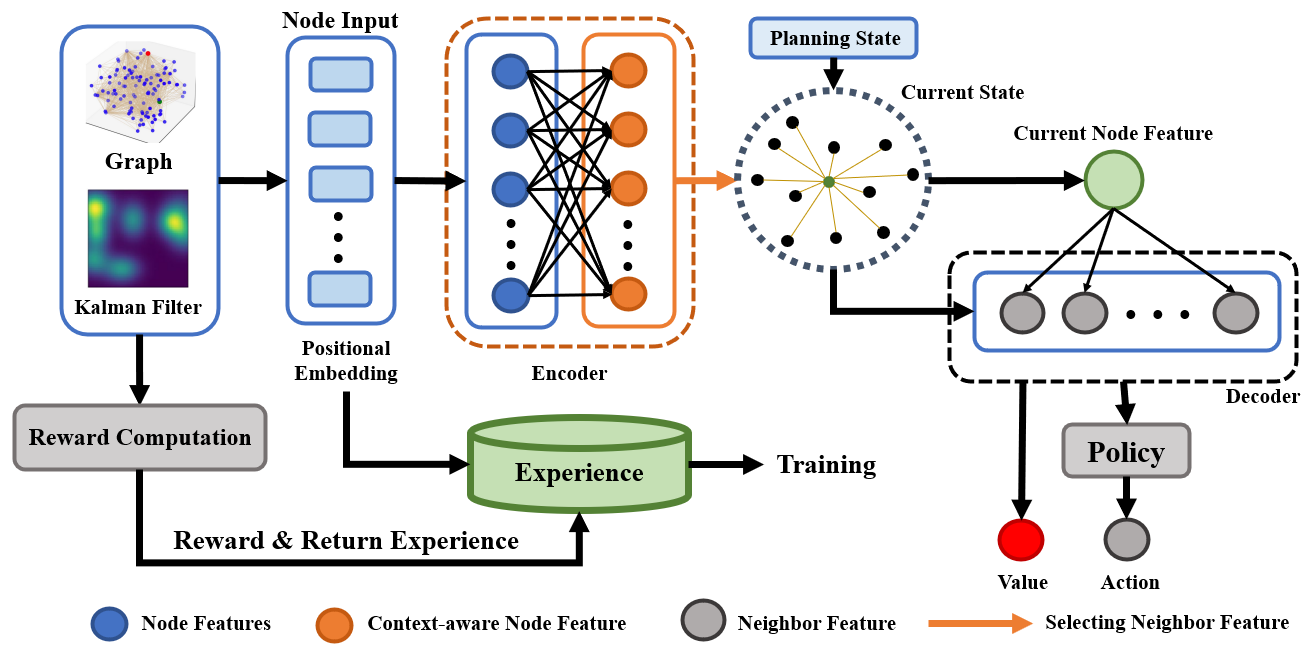}
    \caption{\textbf{Our proposed attention-based neural network for 3D IPP.} The encoder module employs a self-attention layer to capture global dependencies between nodes within the agent’s belief (i.e., the augmented graph) as \emph{context-aware node features}. Leveraging both the current and neighboring context-aware node features (illustrated by the gray dashed circle), our decoder uses a ``pointer network'' structure with cross-attention, allowing the model to naturally scale to different numbers of neighbors, and generates a context-aware policy along with a value estimate during training.}
    \label{F4}
    \vspace{-0.2cm}
\end{figure*}

\subsubsection{Reward Structure} 
To encourage efficient exploration and maximize the information gain, the agent is rewarded for reducing uncertainty in the identified regions of interest, \(\mathcal{X}_I\), when taking measurements. The reward function is defined as:
\begin{equation} 
r_t = \left(\frac{\mathrm{Tr}(\mathbf{P^-}) - \mathrm{Tr}(\mathbf{P^+})}{\mathrm{Tr}(\mathbf{P^-})}\right) \times 10,
\end{equation}
where \(\mathrm{Tr}(\mathbf{P^-})\) and \(\mathrm{Tr}(\mathbf{P^+})\) represent the total uncertainty within \(\mathcal{X}_I\) before and after the measurement, respectively. To stabilize training, the reward is normalized, and a scaling factor of 10 is applied to prevent vanishing gradients.

\subsubsection{Network Architecture} 
Our proposed attention-based neural network is designed to facilitate adaptive exploration in the constructed environment. The architecture, illustrated in Fig.~\ref{F4}, is inspired by CAtNIPP~\cite{cao2023catnipp}. The model comprises an \emph{encoder}, which captures spatial dependencies among nodes in the augmented graph \(G'\), and a \emph{decoder}, which generates action policies based on the extracted features and the agent's current state \(\{ G', v_c, s_c \}\).

Since the graph state \(G'\) is inherently incomplete—each node is only connected to a fixed number of neighbors—the encoder employs a Transformer attention layer with \emph{graph positional encoding} (PE) based on Laplacian eigenvectors~\cite{dwivedi2020benchmarking}. This encoding enhances the network’s ability to recognize structural connectivity within the graph. Additionally, to accommodate a dynamically changing action space, the decoder incorporates a \emph{Pointer Network}~\cite{vinyals2015pointer}, allowing the model to flexibly adapt to different numbers of neighboring nodes.


\subsubsection{Attention Layer}
The core of our model is the Transformer-based attention mechanism~\cite{vaswani2017attention}, which efficiently aggregates information across spatially distributed observations. The attention mechanism operates on query (\(h^q\)) and key-value (\(h^{k,v}\)) sources, both represented as feature vectors of the same dimension. The attention weights are computed based on the similarity between query and key vectors:
\begin{equation}
\begin{aligned}
q_i &= W^Q h^q_i, \quad k_i = W^K h^{k,v}_i, \quad v_i = W^V h^{k,v}_i, \\
u_{i,j} &= \frac{q_i^\top k_j}{\sqrt{d}}, \quad a_{i,j} = \frac{e^{u_{i,j}}}{\sum_{j=1}^{n} e^{u_{i,j}}}, \quad
h'_i = \sum_{j=1}^{n} a_{i,j} v_j.
\end{aligned}
\end{equation}
Here, \(W^Q\), \(W^K\), and \(W^V\) are learnable weight matrices of size \(d \times d\). The updated features are further processed by a feedforward sublayer consisting of two linear transformations with an intermediate ReLU activation. As in~\cite{vaswani2017attention}, layer normalization and residual connections are applied to stabilize training.

\subsubsection{Encoder}
The encoder models the environment by capturing dependencies between nodes in the augmented graph \(G'\). Initially, each node \(v'_i \in V'\) is embedded into a \(d\)-dimensional feature vector \(h^n_i\), incorporating both positional and structural information:
\begin{equation} 
h^n_i =
\begin{cases}
W^{L} v'_i + b^L + W^{PE} \lambda_i + b^{PE}, & i > 0, \\
W^D v'_0 + b^D + W^{PE} \lambda_0 + b^{PE}, & i = 0.
\end{cases}
\end{equation}

Here, \(\lambda_i\) represents the precomputed \(k\)-dimensional Laplacian eigenvector encoding the node’s position in the graph, while \(W^L, W^D, W^{PE}, b^L, b^D,\) and \(b^{PE}\) are learnable parameters. The node embeddings are then processed by an attention layer, where \(h^q = h^{k,v} = h^n\), to generate the final context-aware node representations \(h^{en}_i\), which incorporate dependencies between each node and its neighbors.

\begin{table*}[t]
\small
\tabcolsep=0.23cm
\renewcommand\arraystretch{1.2}
\centering
\caption{\textbf{Performance evaluation of different IPP methods.}}
\resizebox{\textwidth}{!}{  
\begin{tabular}{c|c|cccc|cccc|c}
\toprule
\multirow{2}{*}{\textbf{Environment}} & \multirow{2}{*}{\textbf{Method}} & \multicolumn{4}{c|}{\textbf{Uncertainty Reduction}} & \multicolumn{4}{c|}{\textbf{RMSE Reduction}} & \multirow{2}{*}{\textbf{Runtime}} \\ 
\cline{3-10}
& & 50s & 100s & 150s & 200s & 50s & 100s & 150s & 200s & \\ 
\midrule
\multirow{4}{*}{\textbf{Offline Evaluation}} &\textbf{Ours} & \textbf{51.72}\% & \textbf{72.06}\% & \textbf{84.69}\% & \textbf{86.43}\% & \textbf{38.76}\% & \textbf{53.52}\% & \textbf{61.80}\% & \textbf{61.52}\% & \textbf{0.0054s} \\ 
& MCTS & 34.09\% & 53.51\% & 61.81\% & 72.67\% & 36.39\% & 51.80\% & 59.08\% & 58.53\% & 3.9397s \\ 
& Random & 30.23\% & 43.15\% & 58.31\% & 62.65\% & 15.74\% & 33.19\% & 41.97\% & 37.45\% & 0.0001s \\ 
& Coverage & 22.11\% & 34.92\% & 51.54\% & 56.39\% & 5.28\% & 19.67\% & 33.26\% & 33.08\% & 0.0001s \\ 
\midrule
\multirow{2}{*}{\textbf{Simulated Deployment}} & \textbf{\textbf{Ours}} & \textbf{47.70}\% & \textbf{70.06}\% & \textbf{74.65}\% & \textbf{80.18}\% & \textbf{35.76}\% & \textbf{54.42}\% & \textbf{59.22}\% & \textbf{67.29}\% & \textbf{0.0197s} \\
& CMA-ES & 42.19\% & 64.20\% & 68.43\% & 76.92\% & 33.44\% & 48.89\% & 51.27\% & 64.83\% & 0.0218s \\
\bottomrule
\end{tabular}
} 
\label{compar}
\vspace{-0.5cm}
\end{table*}

\subsubsection{Decoder}
The decoder generates the agent’s policy based on the context-aware node features. Given the agent’s current position \(v_c\) and the edge set \(E\), we extract the current node feature \(h^c\) and its connected neighbors \(h^n\). These features are passed through an attention layer, in which $h^q = h^{ec},\quad h^{k,v} = h^n$. The output, \(\hat{h}^{ec}\), is then used in two ways: (1) it is passed through a linear layer to compute the state value function \(V(s_t)\), and (2) it serves as input to a second attention layer, where $h^q = \hat{h}^{ec},\quad h^{k,v} = h^n$. This final attention operation produces an action representation \(u_i\), which is subsequently normalized into a probability distribution \(\pi\) for selecting the next node to visit. This architecture enables the model to dynamically adapt to different graph structures, making it well-suited for large-scale 3D exploration tasks.

\subsection{RL Training}

After constructing the attention-based neural network, we employ the Proximal Policy Optimization (PPO) algorithm~\cite{schulman2017proximal} to train the agent. PPO provides a balance between stability and efficiency by incorporating the advantages of trust region methods while reducing sample complexity. The objective function for PPO is formulated as follows:
\begin{small}
\begin{equation} 
L^{CLIP}(\theta)=\hat{\mathbb{E}}_t \Big[ \min \big( r_t(\theta) \hat{A}_t, \mathrm{clip}(r_t(\theta), 1-\epsilon, 1+\epsilon) \hat{A}_t \big) \Big],
\end{equation}
\end{small}
where $r_t(\theta)=\frac{p_\theta(\phi_{t}|s_{t})}{p_{\theta_{old}}(\phi_t|s_t)}$ represents the probability ratio between the updated and previous policies. The clipping mechanism prevents excessively large updates, thereby stabilizing training and improving convergence.

For training, carefully chosen hyperparameters and optimization settings are applied, including the Adam optimizer. The robot's initial belief is uniformly distributed across the environment, and specific environmental variables and training parameters are detailed as follows. Each training episode generates a randomized ground truth map, where grid cells are independently assigned Bernoulli-distributed values representing target occupancy probabilities~\cite{elfes1989using}. To regulate measurement frequency, new measurements are taken whenever the agent travels a distance exceeding $0.2$ units since the last recorded measurement.

The PPO algorithm runs for eight iterations per training episode, with a Gaussian Process prior $\mathcal{GP}(\mathbf{0,1})$, with $k=20$ neighboring nodes, a mission budget of $150\rm s$, a batch size of $128$, a learning rate at $10^{-5}$, and a discount factor $\gamma = 0.99$. One-step TD-error is used for advantage estimation. Each training episode consists of eight PPO iterations. Training is conducted on an AMD Ryzen 7 5800X CPU and an NVIDIA GeForce RTX 3060 GPU, with multiprocessing enabled to run five IPP instances in parallel. This parallelization strategy significantly accelerates data collection and improves model convergence. The entire training process requires approximately $10$ hours to reach a stable policy.

\section{Evaluation}
\label{sec:eval}

This section describes our testing procedure and presents our experimental results. We first evaluate the proposed RL approach against baseline methods under the same environmental conditions used during training. We then examine its generalization capability by testing on monitoring maps of varying sizes. Finally, we validate its practical feasibility in real-world scenarios through Gazebo-based simulations for weed detection tasks.

\subsection{Experimental Setup}
Before the training and experiment, the specific environmental settings are established. The simulation setup considers terrains $\xi$ with 2D discrete field maps $\mathcal{X}$ with values between 0 and 1, randomly split into high- and low-value regions to define regions of interest. There is a set of discretized nodes in 3D space includes two altitude levels ($8\rm m$ and $14\rm m$).

Camera measurement noise is simulated using an altitude-dependent inverse sensor model from Popovic \emph{et al}.~\cite{ruckin2022adaptive}, assuming a downward-facing square camera with $60^{\circ}$ field of view. The prior map is normalized first with mean value at $0.5$, The GP is defined with length scale $3.67$, signal variance $1.82$, and noise variance $1.42$ by maximizing log marginal likelihood over independent maps. Regions of interest are identified using a threshold $\mu_{th} = 0.4$. The mission budget $B=150\rm s$, with the UAV starting at  $(2,2,14)$, with maximum speed $u_v=2\rm m/s$. 

During training, each episode generates a randomized ground truth map, and UAV start positions to mitigate overfitting. Planning performance is evaluated based on two key metrics: map uncertainty \(\mathrm{Tr}(\mathbf{P^+})\) and root mean squared error (RMSE) within regions of interest, where lower values indicate improved performance. Unlike previous works~\cite{popovic2020informative, hitz2017adaptive, choudhury2020adaptive}, our evaluation considers both path travel time and planning runtime, ensuring a realistic assessment of computational constraints faced by robotic platforms with limited onboard resources. The effective mission time is defined as the budget consumed throughout the task execution.

\begin{figure*}[th]
  \centering
    \includegraphics[width=.25\textwidth]{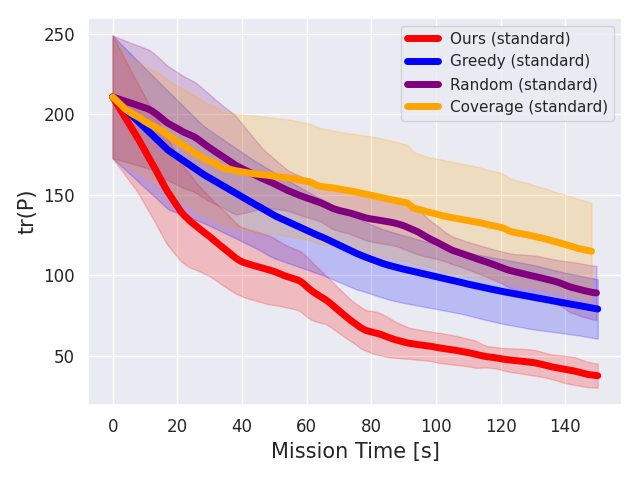}
    \includegraphics[width=.25\textwidth]{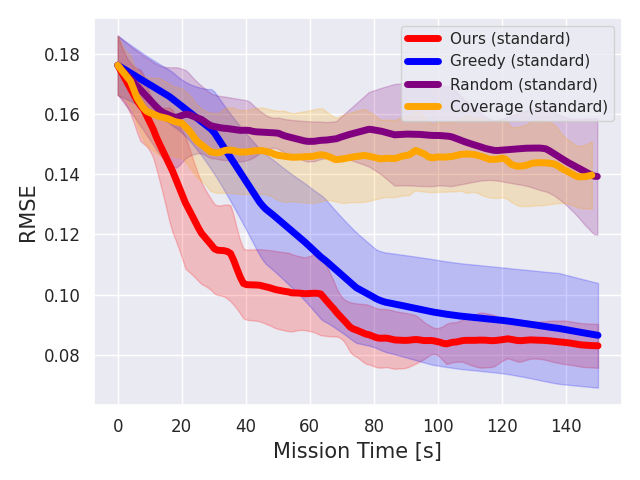}
    \includegraphics[width=.235\textwidth]{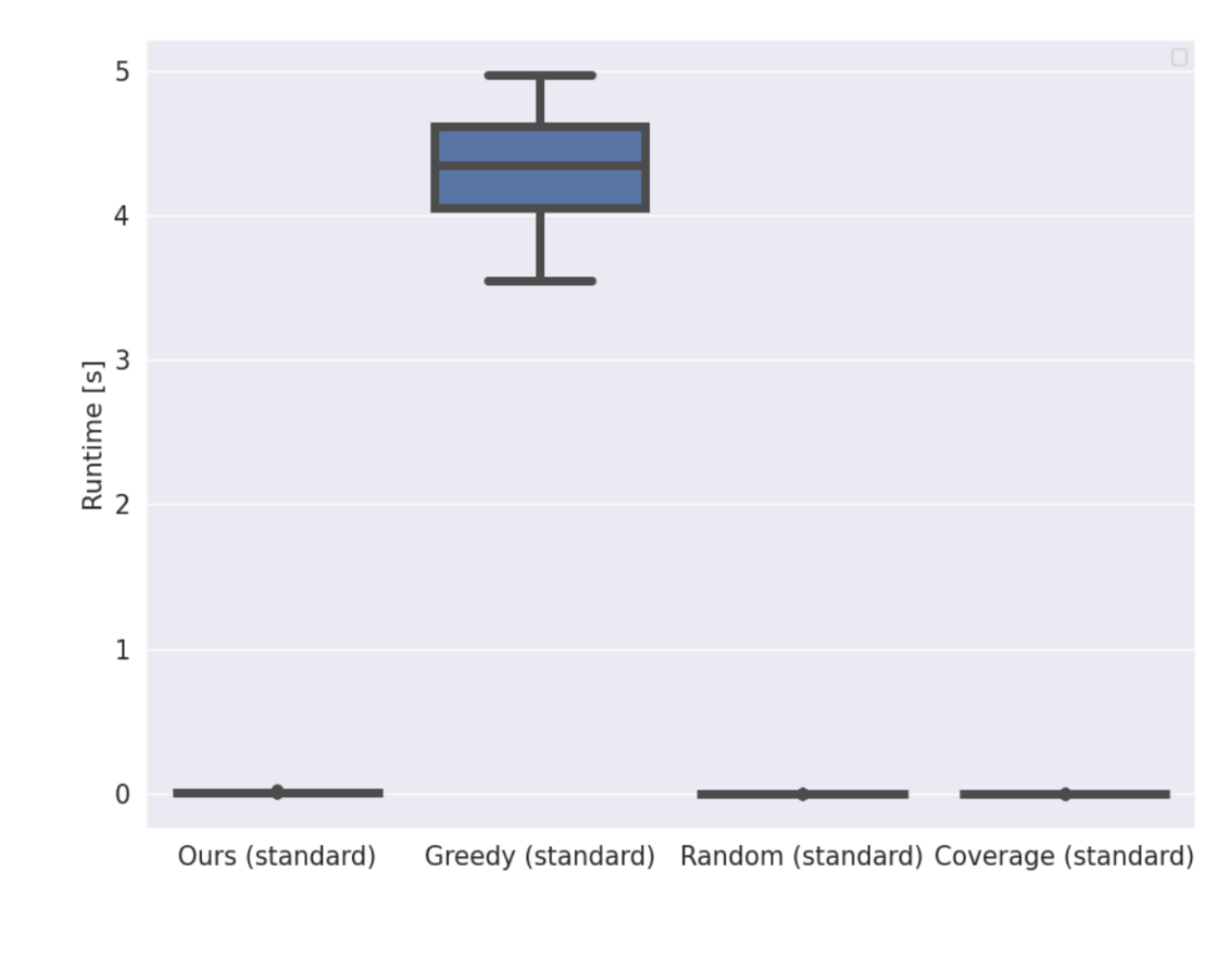}
    \includegraphics[width=.235\textwidth]{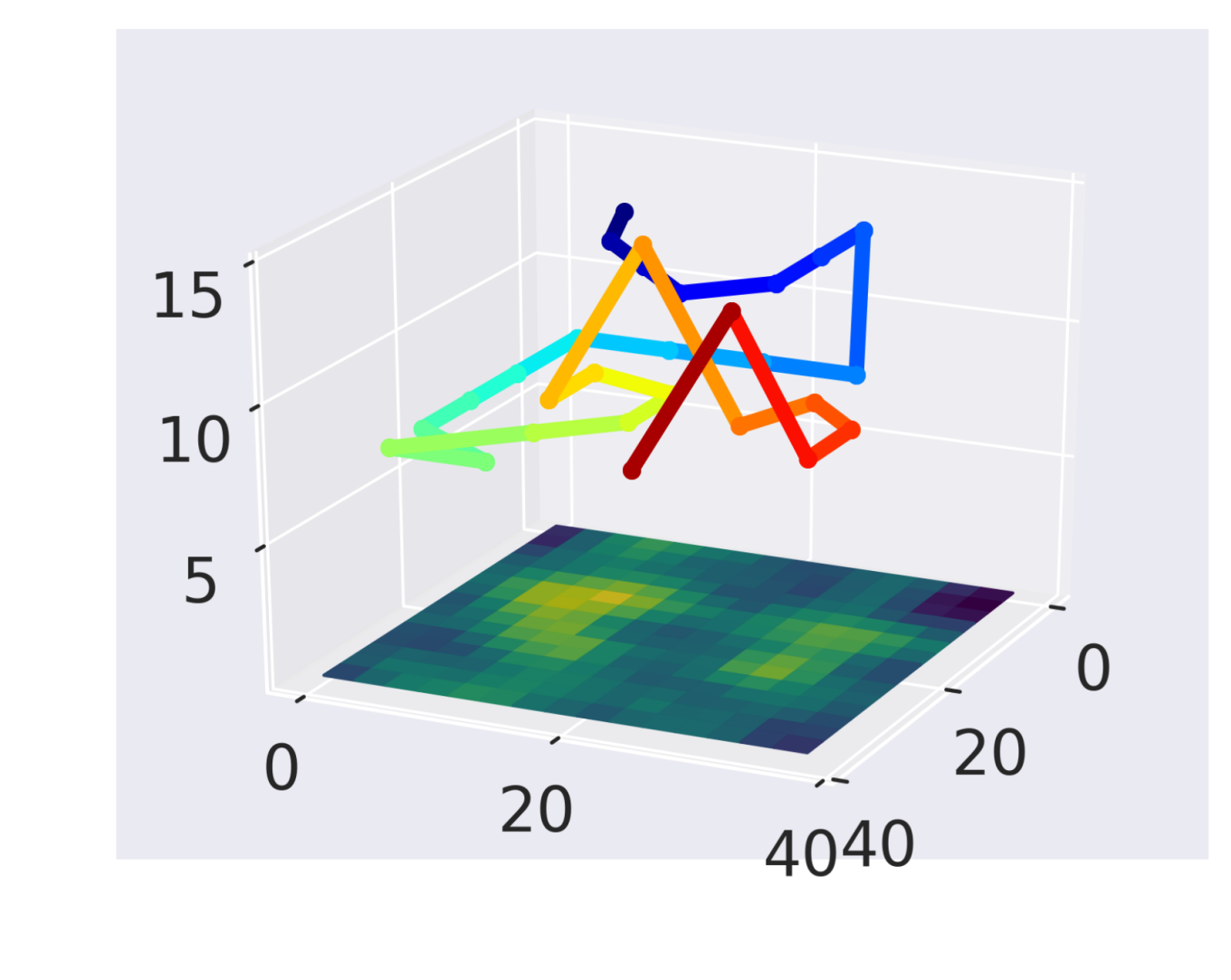}
    \vspace{-0.3cm}
    \caption{\textbf{Evaluation of our approach against baselines.} On average, our RL approach ensures the fastest uncertainty (left) and RMSE (center left) reduction in regions of interest over mission time. Solid lines indicate means over 4 trials, and shaded regions associated standard deviations. Note that the randomly generated information hotspots may influence the performance of agent. Our method guarantees consistent performance across various maps while substantially reducing replanning runtime (center right). Right: The evolving path (transitioning from blue to red over time) demonstrates the adaptive capability of our approach, effectively exploring the terrain with a focus on high-value regions.}
    \label{result}
    \vspace{-0.4cm}
\end{figure*}

\subsection{Comparison Against Baselines}

Next, our RL algorithm is evaluated against baselines. We set a resolution $r=2.5\rm m$ and a grid map $\mathcal{X}$ is $15\times15$, hence $\mathcal{A}$ has $450$ actions. Our approach is compared against: (a) uniform random sampling in $\mathcal{A}$; (b) coverage path with equispaced measurements at a fixed $8\rm m$ altitude; (c) Monte Carlo tree search (MCTS) with progressive widening~\cite{sunberg2018online} for large action spaces and a generalized cost-benefit rollout policy proposed by Choudhury \emph{et al}.~\cite{choudhury2020adaptive}. Considering that~\cite{ruckin2022adaptive} focuses more on planning future measurement points in advance using MCTS in 2D environments, whereas our work emphasizes real-time adaptive decision-making in 3D space, it was not included in the experiments. Our approach demonstrated superior performance under all tested budgets and achieved a relatively lower running time than prior works, even though the agent was trained on the discretized action space~\cite{popovic2020informative}.

Figure~\ref{result} presents the results obtained using each approach. As shown in the trajectory, the agent initially takes measurements at \(14\rm m\) to reduce overall uncertainty before descending to \(8\rm m\) to refine high-interest areas. Our method significantly reduces runtime, achieving a tenfold speedup compared to MCTS and other baseline approaches.


Our model achieves the fastest reduction in both uncertainty and RMSE, with uncertainty dropping below 50 within approximately 100 seconds, outperforming other methods. Random sampling performs poorly, as it only reduces uncertainty and RMSE in high-value regions by chance. The coverage method shows high variability, depending on hotspot alignment with its trajectory. Experimental results in Table~\ref{compar} demonstrate that the attention-based model consistently outperforms baselines in reliability and efficiency.

\subsection{Generalization Capacity}

As shown in previous work~\cite{vaswani2017attention}, attention mechanisms are well-suited for handling inputs of varying sizes. This property enables our trained attention-based model to naturally adapt to maps of different dimensions without requiring retraining. To assess its generalization capability, we evaluated the model on two grid sizes: \(15\times15\) (matching training environments) and \(20\times20\) (larger environment).

Our results demonstrate that our model consistently achieves optimal performance across different map sizes, whereas other approaches require retraining when the grid size changes. This adaptability underscores the model’s robustness in varying environments and its potential for real-world deployment without frequent retraining.
\begin{figure}[H]
  \centering
    \includegraphics[width=.235\textwidth]{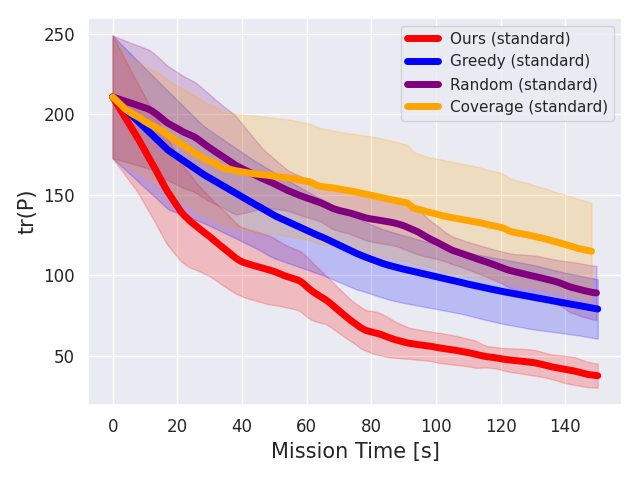}
    \includegraphics[width=.235\textwidth]{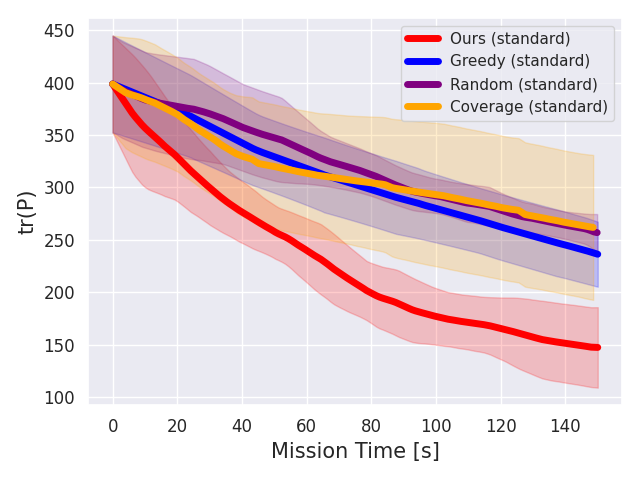}
    \caption{\textbf{Test results in $15\times15$ and $20\times20$ environments}. Although our agent was only trained on $15\times15$ size maps, it outperforms all baselines on this larger map.}
    \label{diff_size}
    \vspace{-0.2cm}
\end{figure}

\subsection{Gazebo-based Simulation}

\begin{figure}[t]
  \centering
    \includegraphics[width=.2\textwidth]{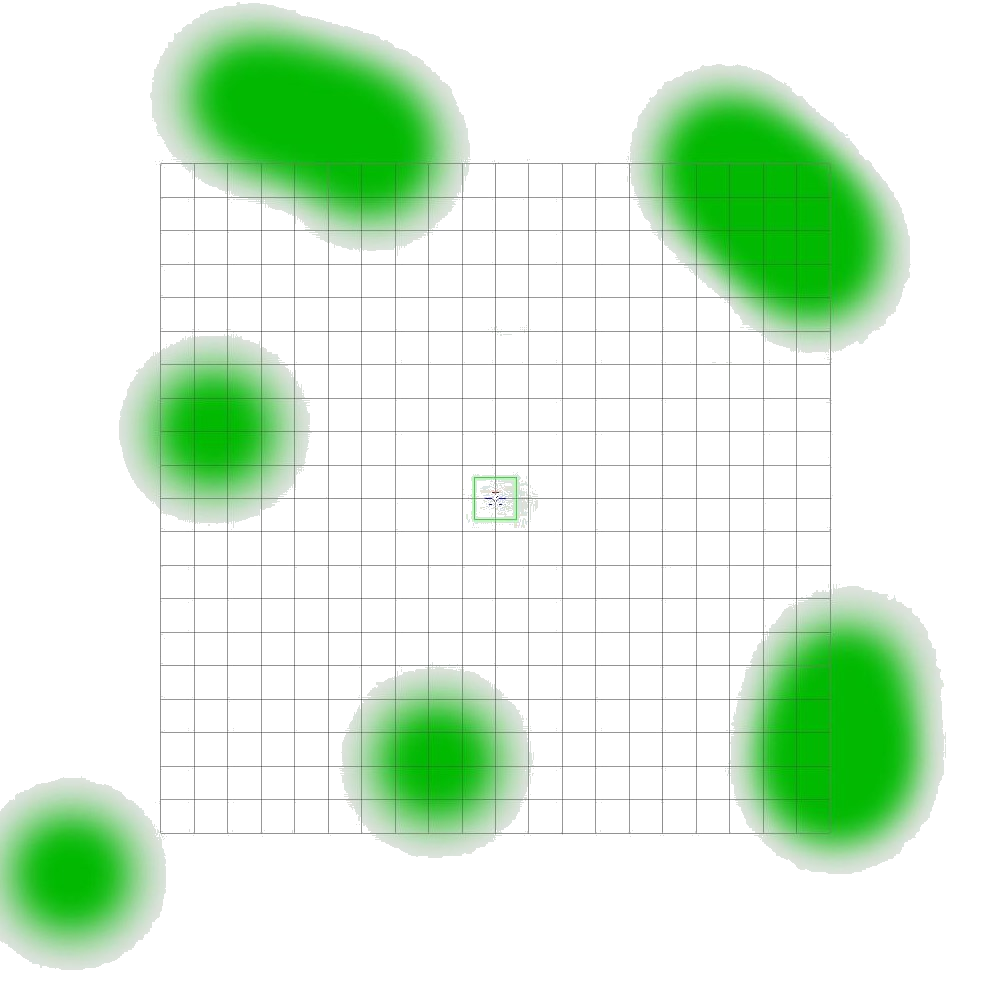}
    \includegraphics[width=.265\textwidth]{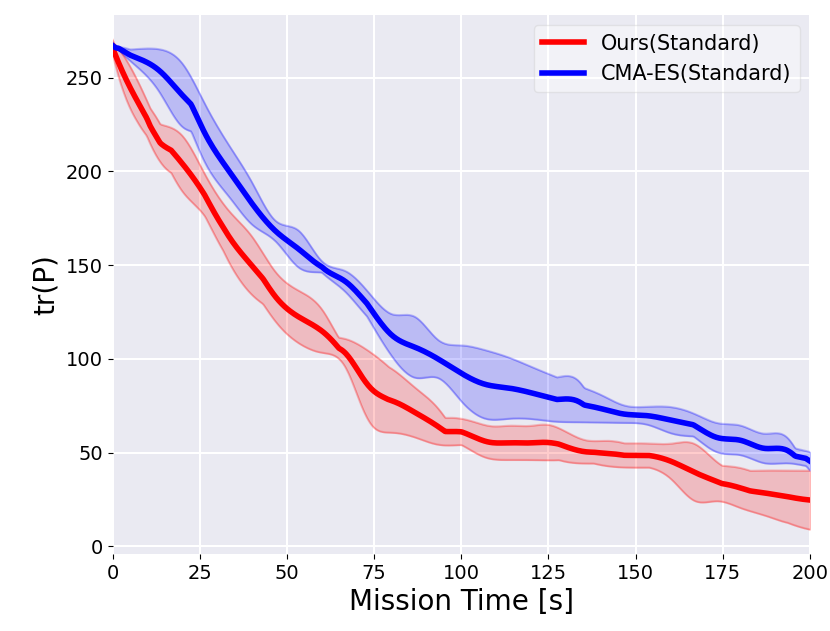}
    \caption{\textbf{Simulation environment and performance comparison.} \emph{(Left)} Top-down view of the simulated environment in Gazebo, where green regions indicate weed distributions. \emph{(Right)} Trace of the covariance matrix $\text{tr}(P)$ over mission time on $15\times15$ maps. Our method achieves faster convergence and lower uncertainty throughout the mission.}
    \label{ros_env}
    \vspace{-0.6cm}
\end{figure}

We demonstrate our approach through a Gazebo-based simulation in a weed detection scenario. The simulation environment (Fig. \ref{ros_env}-left) comprises a $37.5m\times37.5m$ map, which can be discretized into a $15\times15$ grid map with a resolution of $r=2.5m$. Regions of interest are identified using a threshold $\mu_{th} = 0.05$. The mission parameters include a budget of $B=200\rm s$ and a UAV maximum speed of $u_v=2\rm m/s$. The UAV begins by navigating to the initial position $(2,2,14)$, after which the experiment commences and the budget countdown initiates. As illustrated in Fig.\ref{ros_env}-right, our approach achieves significantly faster uncertainty reduction compared to CMA-ES, underscoring its suitability for real-time, adaptive decision-making in dynamic environments.

\section{Conclusion and Future Work}
\label{sec:conclusion}

In this work, we propose an attention-based neural network for 3D adaptive informative path planning under time constraints, enabling efficient information collection in active sensing. Our approach integrates altitude-aware decision-making, balancing flight altitude, sensor footprint, and measurement accuracy. Leveraging self-attention, the model builds a global context representation, guiding non-myopic, sequential decisions to optimize short- and long-term objectives.
Our experimental results demonstrate that our approach significantly outperforms state-of-the-art methods in reducing environmental uncertainty and reconstructing terrain features with greater accuracy, particularly in value-dependent regions of interest. Our model not only achieves faster uncertainty reduction but also generalizes well to varying environment sizes, highlighting its scalability and robustness.

Future work will focus on enhancing the efficiency of our approach for large-scale field trials, validating its real-world applicability under more complex and dynamic environmental conditions. We also plan to extend our method to multi-agent scenarios, where multiple aerial robots collaboratively explore and gather information while dynamically coordinating their paths. Furthermore, we aim to integrate uncertainty-aware exploration strategies that explicitly balance exploitation and exploration over long horizons, improving robustness in highly uncertain or dynamically changing environments. Finally, we will investigate incorporating more complex sensor models, such as multi-modal fusion, to enhance adaptability in diverse real-world applications.

\bibliographystyle{unsrt}
\bibliography{bib}

\addtolength{\textheight}{-12cm}   

\end{document}